\documentclass[11pt,a4paper]{article}
\usepackage[hyperref]{acl2020}
\usepackage{times}
\usepackage{latexsym}

\usepackage{microtype}

\usepackage{csquotes}
\usepackage{xspace}
\usepackage{amsmath}
\usepackage{amssymb}
\usepackage{graphicx}
\usepackage{upgreek}
\usepackage{booktabs}
\usepackage{lipsum}
\usepackage{url}
\usepackage{xcolor}
\usepackage{microtype}

\newcommand{\code}[1]{\texttt{#1}}
\newcommand{\cmd}[1]{\textbf{\small{\code{#1}}}}
\newcommand{\scmd}[1]{\textbf{\scriptsize{\code{#1}}}}

\newcounter{lintr}

\newcommand{\smx}{{\mathrm{softmax}}}

\newcommand{\relu}{{\mathrm{ReLU}}}
\newcommand{\mlp}{{\mathrm{MLP}}}

\newcommand{\qadqn}{QA-DQN\xspace}

\newcommand{\dqn}{DQN\xspace}
\newcommand{\aac}{A2C\xspace}

\newcommand{\fone}{$\text{F}_1$\xspace}
\newcommand{\foneinfo}{$\text{F}_{1\text{info}}$\xspace}

\definecolor{green1}{HTML}{009A36}
\definecolor{blue1}{HTML}{2933B0}
\definecolor{purple1}{HTML}{880AAD}

\newcommand{\isquad}{iSQuAD\xspace}
\newcommand{\inewsqa}{iNewsQA\xspace}
\newcommand{\imrc}{iMRC\xspace}
\newcommand{\squad}{SQuAD\xspace}
\newcommand{\newsqa}{NewsQA\xspace}
\newcommand{\ctrlf}{Ctrl+F\xspace}
\newcommand{\query}{\textcolor{purple1}{\small{QUERY}}\xspace}

\aclfinalcopy % Uncomment this line for the final submission
\title{Interactive Machine Comprehension with Information Seeking Agents}

\author{Xingdi Yuan$^\dag$\thanks{\:\:\:\:Equal contribution.} \:\:\:\: Jie Fu$^{\ddag\spadesuit}$\footnotemark[1] \:\:\:\: Marc-Alexandre C\^ot\'{e}$^\dag$ \:\:\:\: Yi Tay$^{\diamondsuit}$ \\ 
\textbf{Christopher Pal$^{\ddag\spadesuit}$ \:\:\:\: Adam Trischler$^{\dag}$}\\
$^\dag$Microsoft Research, Montr\'{e}al \:\:\:\: $^\ddag$Polytechnique Montr\'{e}al \\ $^\spadesuit$ Mila \:\:\:\: $^\diamondsuit$ Nanyang Technological University\\
eric.yuan@microsoft.com \:\:\:\: jie.fu@polymtl.ca
}

\date{}

\begin{document}
\maketitle
\begin{abstract}
Existing machine reading comprehension (MRC) models do not scale effectively to real-world applications like web-level information retrieval and question answering (QA). We argue that this stems from the nature of MRC datasets: most of these are \textit{static} environments 
% where models have access to the entire collection of supporting documents to answer the questions.
wherein the supporting documents and all necessary information are fully observed.
In this paper, we propose a simple method that reframes existing MRC datasets as \textit{interactive}, partially observable environments.
Specifically, we ``occlude'' the majority of a document's text and add context-sensitive commands that reveal ``glimpses'' of the hidden text to a model.
We repurpose \squad and \newsqa as an initial case study, and then show how the interactive corpora can be used to train a model that seeks relevant information through sequential decision making.
We believe that this setting can contribute in scaling models to web-level QA scenarios.\footnote{The dataset and implementation of our baseline agents are publicly available at \url{https://github.com/xingdi-eric-yuan/imrc_public}. }
\end{abstract}

\section{Introduction}
\label{section:intro}

Many machine reading comprehension (MRC) datasets have been released in recent years \citep{rajpurkar16squad,trischler16newsqa,nguyen16msmarco,reddy18coqa,yang18hotpot} to benchmark a system's ability to understand and reason over natural language. Typically, these datasets require an MRC model to read through a document to answer a question about information contained therein.

The supporting document is, more often than not, static and fully observable. This raises concerns, since models may find answers simply through shallow pattern matching; e.g., syntactic similarity between the words in questions and documents. As pointed out by \citet{sugawara18easier}, for questions starting with \textit{when}, models tend to predict the \textbf{only} date/time answer in the supporting document.
%Similarly, this might also be the case when predicting entities based on the \textit{who} type questions.
Such behavior limits the generality and usefulness of MRC models, and suggests that they do not learn a proper `understanding' of the intended task.
In this paper, to address this problem, we shift the focus of MRC data away from `spoon-feeding' models with sufficient information in fully observable, static documents. Instead, we propose \textit{interactive} versions of existing MRC tasks, whereby the information needed to answer a question must be gathered sequentially.

\begin{table}[t!]
    \scriptsize
    \centering
    \begin{tabular}{r|l|r}
        \toprule
        \multicolumn{3}{c}{\textcolor{red}{Question: What was the Harvard endowment total in 2011 ?}} \\
        \midrule
        $o_1$               &  \multicolumn{2}{l}{Harvard has the largest university endowment in the world .} \\
        \midrule
        \multicolumn{2}{r|}{\textcolor{white}{WWWWWWWWWWWWWWWWWWWWwwwww}\scmd{next}} & $a_1$\\
        \midrule
        $o_2$               &  \multicolumn{2}{l}{At the end of June 2009, it was worth \$25.7 billion, about 30\%}\\
                            &  \multicolumn{2}{l}{less than at the same time in 2008.} \\
        \midrule
        \multicolumn{2}{r|}{\scmd{\ctrlf Harvard}} & $a_2$\\
        \midrule
        $o_3$               &  \multicolumn{2}{l}{In December 2008, \underline{Harvard} announced that its endowment had}\\
                            &  \multicolumn{2}{l}{lost 22\% from July to October 2008, necessitating budget cuts.} \\
        \midrule
        \multicolumn{2}{r|}{\scmd{\ctrlf 2011}} & $a_3$\\
        \midrule
        $o_4$               &  \multicolumn{2}{l}{As of September \underline{2011} , it had nearly regained the loss suffered}\\
                            &  \multicolumn{2}{l}{during the 2008 recession .} \\
        \midrule
        \multicolumn{2}{r|}{\scmd{\ctrlf 2011}} & $a_4$\\
        \midrule
        $o_5$               &  \multicolumn{2}{l}{It was worth \textbf{\textcolor{red}{\$ 32}} \textbf{\textcolor{red}{billion}} in \underline{2011} , up from \$ 28 billion in}\\
                            &  \multicolumn{2}{l}{September 2010 and \$ 26 billion in 2009 .} \\
        \midrule
        \multicolumn{2}{r|}{\scmd{stop}} & $a_5$\\
        \midrule
        \multicolumn{3}{c}{\textcolor{red}{Prediction: \scmd{\$ 32 billion}}} \\
        \bottomrule
    \end{tabular}
    \caption{Example of the interactive machine reading comprehension behavior.}
    \label{tab:cherry_pick_astract}
\end{table}

The key idea behind our proposed interactive MRC (\imrc) is to restrict the document context that a model observes at one time.
Concretely, we split a supporting document into its component sentences and withhold these sentences from the model.
Given a question, the model must issue commands to observe sentences in the withheld set; we equip models with actions such as \cmd{\ctrlf} to search for matches to a \query within \textit{partially} observed documents.
A model searches iteratively, conditioning each command on the input question and the sentences it has observed previously.
Thus, our task requires models to `feed themselves' rather than spoon-feeding them with information.
This casts MRC as a sequential decision-making problem amenable to reinforcement learning (RL).

Our proposed approach % is orthogonal with existing MRC dataset and models, it 
lies outside of traditional QA work, the idea 
can be applied to almost all existing MRC datasets and models to study interactive information-seeking behavior.
As a case study in this paper, we re-purpose two well known, related corpora with different difficulty levels for our \imrc task: \squad and \newsqa.
Table~\ref{tab:cherry_pick_astract} shows an example of a model performing interactive MRC on these datasets.
Naturally, our reframing makes the MRC problem harder; however, we believe the added demands of \imrc more closely match web-level QA and may lead to deeper comprehension of documents' content.

The main contributions of this work are as follows:
\begin{enumerate}
    \item We describe a method to make MRC datasets interactive and formulate the new task as an RL problem.
    \item We develop a baseline agent that combines a top performing MRC model and two state-of-the-art RL optimization algorithms and test it on \imrc tasks.
    \item We conduct experiments on several variants of \imrc and discuss the significant challenges posed by our setting.
\end{enumerate}
% \textbf{1.} We describe a method to make MRC datasets interactive and formulate the new task as an RL problem.\\
% \textbf{2.} We develop a baseline agent that combines a top performing MRC model and two state-of-the-art RL optimization algorithms and test it on \imrc tasks. \\
% \textbf{3.} We conduct experiments on several variants of \imrc and discuss the significant challenges posed by our setting.
% \begin{enumerate}
%     \item We describe a method to make MRC datasets interactive and formulate the new task as an RL problem.
%     \item We develop a baseline agent that combines a top performing MRC model and a state-of-the-art RL optimization algorithm and test it on our \imrc tasks. 
%     \item We conduct experiments on several variants of \imrc and discuss the significant challenges posed by our setting.
% \end{enumerate}

\section{Related Works}
\label{section:related}
% \subsection{Streaming setting}

% \subsection{Skip Reading and Re-Reading}
Skip-reading \citep{Yu2017,Seo2017,Choi2017} is an existing setting in which MRC models read partial documents. 
Concretely, these methods assume that not all tokens in the input sequence are equally useful, and therefore learn to skip irrelevant tokens. % based on the current input and their internal memory.
Since skipping decisions are discrete, the models are often optimized by the REINFORCE algorithm \citep{williams1992reinforce}. 
For example, the structural-jump-LSTM \citep{Hansen2019} learns to skip and jump over chunks of text, whereas \citet{Han2019} designed a QA task where the model reads streaming data without knowing when the question will be provided. 
Skip-reading approaches are limited in that they only consider jumping \textit{forward} over a few consecutive tokens. 
Based on the assumption that a single pass of reading may not provide sufficient information, multi-pass reading methods have also been studied \citep{Sha2017,Shen2017}. 
% Our approach is also related to exploring optimal stopping in the setting of reading comprehension \cite{Shen2017}, where an RL agent needs to decide when to stop reading. 

Compared to skip-reading and multi-pass reading, our work enables an agent to jump through a document in a more dynamic manner, in some sense combining aspects of skip-reading and re-reading.
Specifically, an agent can choose to read \textit{forward}, \textit{backward}, or to jump to an \textit{arbitrary position} depending on the query.
This also distinguishes the model we develop in this work from ReasoNet \citep{Shen2017}, a model that decides when to stop \textit{forward} reading.

Recently, there has been various work on and around interactive environments.
For instance, \citet{nogueira16web} proposed WebNav, a tool that automatically transforms a website into a goal-driven web navigation task. They train a neural agent to follow traces using supervised learning. 
\citet{qi2019answering} proposed GoldEn Retriever, an iterative retrieve-and-read system that answers complex open-domain questions, which is also trained with supervised learning.
Although an effective training strategy, supervised learning requires either human labeled or heuristically generated trajectories.
However, there often exist multiple trajectories to solve each question, many of which may not be observed in the supervised data since it is difficult to exhaust all valid trajectories. Generalization can be limited when an agent is trained on such data.

\citet{Bachman2016} introduced a collection of synthetic tasks to train and test information-seeking capabilities in neural models.  
%We extend that work by developing a realistic and challenging text-based task.
\citet{narasimhan16acquiring} proposed an information extraction system that acquires and incorporates external evidence to improve extraction accuracy in domains with limited data. 
\citet{geva2018learning} proposed a DQN-based agent that leverages the (tree) structure of documents and navigates across sentences and paragraphs.
\imrc is distinct from this body of literature in that it does not depend on extra meta information to build tree structures, it is partially-observable, and its action space is as large as 200,000 (much larger than, e.g., the 5 query templates in \citep{narasimhan16acquiring} and tree search in \citep{geva2018learning}).
Our work is also inspired directly by QAit \citep{yuan2019qait}, a set of interactive question answering tasks developed on text-based games.
However, QAit is based on synthetic and templated language which might not require strong language understanding components.
%We extend it to the natural language setting, by leveraging existing MRC tasks which are written in natural language.
This work extends the principle of interactivity to the natural language setting, by leveraging existing MRC tasks already written in natural language.

% Broadly speaking, our approach is also linked to the optimal stopping problem in the literature of Markov decision processes (MDP) \citep{bertsekas2000dp}, where at each game step the agent either continues (without immediate reward) or stops (accumulates reward).
% Here, we reformulate conventional QA tasks through the lens of optimal stopping, in hopes of improving over the shallow matching behaviors exhibited by many MRC systems.
Broadly speaking, our work is also linked to the query reformulation (QR) task in information retrieval literature \citep{nogueira17queryreform}. 
Specifically, QR aims to automatically \textit{rewrite} a query so that it becomes more likely to retrieve relevant documents. 
Our task shares the spirit of iterative interaction between an agent (reformulator in QR) and an environment.
However, the rewritten queries in QR tasks keep the semantic meaning of the original queries, whereas in our task, actions and queries across different game steps can change drastically --- since our task requires an agent to learn a reasoning path (trajectory) towards answering a question, rather than to search the same concept repeatedly.

\section{\imrc: Making MRC Interactive}
\label{section:imrc}

\begin{figure}[t!]
    \centering
    \includegraphics[width=0.4\textwidth]{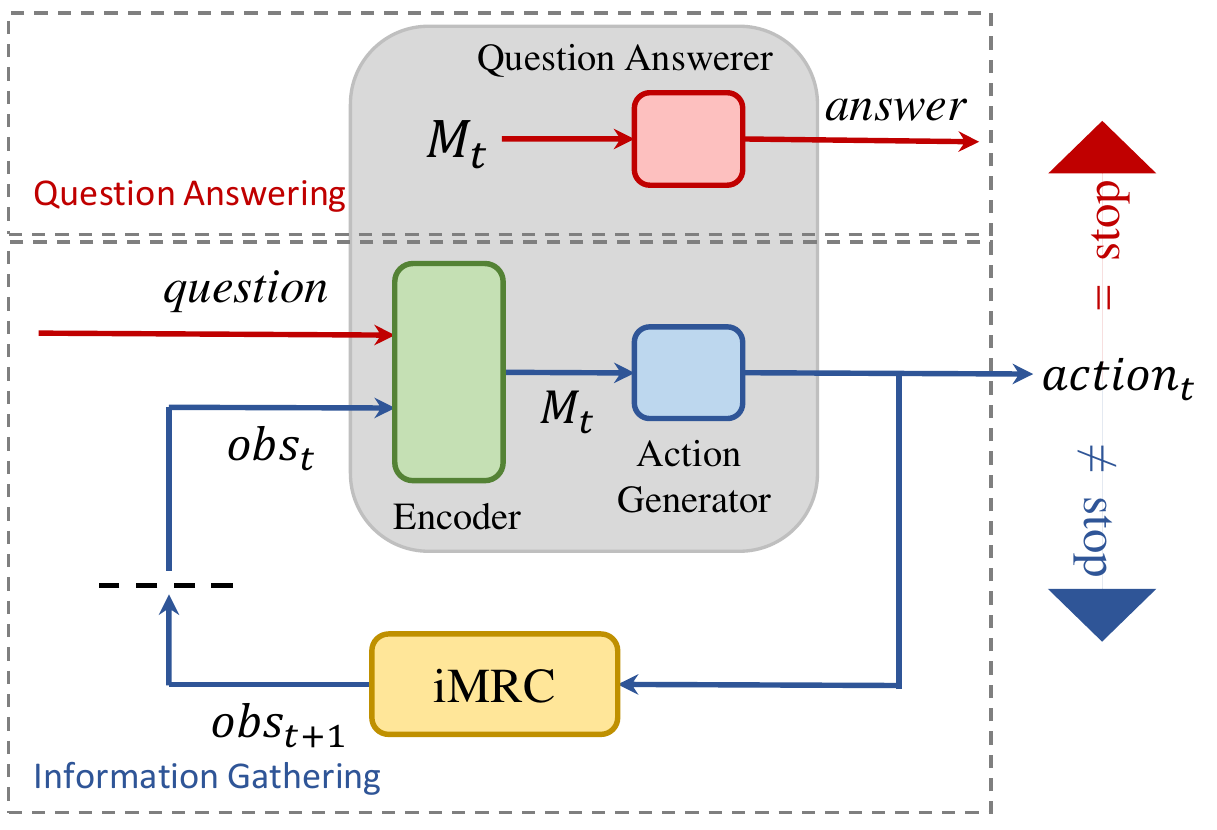}
    \caption{A demonstration of the proposed \imrc pipeline, in which the agent is illustrated as a shaded area. 
    At a game step $t$, it encodes the question and text observation into hidden representations $M_t$. 
    An action generator takes $M_t$ as input to generate commands to interact with the environment.
    If the agent generates \cmd{stop} at this game step, $M_t$ is used to answer question by a question answerer.
    Otherwise, the \imrc environment will provide new text observation in response to the generated action.}
    \label{fig:model}
\end{figure}

The \isquad and \inewsqa datasets are based on \squad v1.1\footnote{We choose \squad v1.1 because in this preliminary study, we focus on extractive question answering.} \citep{rajpurkar16squad} and \newsqa \citep{trischler16newsqa}. 
Both original datasets share similar properties. Specifically, each data-point consists of a tuple, $\{p, q, a\}$, where $p$ represents a paragraph, $q$ a question, and $a$ is the answer. The answer is a word span defined by head and tail positions in $p$. \newsqa is more challenging because it has a larger vocabulary, more difficult questions, and longer source documents.

% We make every data-point an interactable environment, by splitting the paragraph $p$ into sentences, and supporting jump mechanism. 
Every paragraph $p$ is split into a list of sentences $\mathcal{S} = \{s_1, s_2, ..., s_n\}$, where $n$ stands for number of sentences in $p$.
At the start of a question answering episode, an agent observes the question $q$, but rather than observing the entire paragraph $p$, it sees only the first sentence $s_1$ while the rest is withheld.
The agent must issue commands to reveal the hidden sentences progressively and thereby gather the information needed to answer $q$.

The agent should decide when to stop interacting and output an answer, but the number of interaction steps is limited.\footnote{We use 20 as the maximum number of steps, because information revealed by 20 interactions can cover a large portion of the text in either an \isquad or \inewsqa paragraph. A reasonable step budget also speeds up training.} Once the agent has exhausted its step budget, it is forced to answer the question.

\subsection{Interactive MRC as a POMDP}
\label{section:pomdp}
As described in the previous section, we convert MRC tasks into sequential decision-making problems (which we will refer to as \textbf{\textit{games}}). These can be described naturally within the reinforcement learning (RL) framework.
Formally, tasks in \imrc are partially observable Markov decision processes (POMDP)~\citep{kaelbling1998planning}. 
%To act optimally, an agent needs to understand its current observation, and make decision that leads to further observations that are helpful to accomplish the task (i.e., to answer the question correctly).
An \imrc data-point is a discrete-time POMDP defined by $(S, T, A, \Omega, O, R, \gamma)$, where $\gamma \in [0, 1]$ is the discount factor and the other elements are described in detail below.

\textbf{Environment States ($S$):}
The environment state at game step $t$ in the game is $s_t \in S$. 
It contains the environment's underlying conditions (e.g., the semantics and information contained in the document, which part of the document has been revealed so far), much of which is hidden from an agent, the agent can only estimate the state from its partial observations.
% It contains the complete internal information of the game, much of which is hidden from the agent. 
%Note in this work, all environment states remain unchanged.
When the agent issues an action $a_t$, the environment transitions to state $s_{t+1}$ with probability $T(s_{t+1} | s_t, a_t)$. In this work, transition probabilities are either 0 or 1 (i.e., deterministic environment).

\textbf{Actions ($A$):}
At each game step $t$, the agent issues an action $a_t \in A$. We will elaborate on the action space of \imrc in \S~\ref{section:easy_hard_settings} and \S~\ref{section:query}.

\textbf{Observations ($\Omega$):}
The text information perceived by the agent at a given game step $t$ is the agent's observation, $o_t \in \Omega$, which depends on the environment state and the previous action with probability $O(o_t|s_t)$. 
Again, observation probabilities are either 0 or 1 (i.e., noiseless observation).
%Thus, the function $O$ selects from the environment state what information to show to the agent given the last action.

\textbf{Reward Function ($R$):}
Based on its actions, the agent receives rewards $r_t = R(s_t, a_t)$. Its objective is to maximize the expected discounted sum of rewards $E \left[\sum_t \gamma^t r_t \right]$.

\subsection{Easy and Hard Modes}
\label{section:easy_hard_settings}
As a question answering dataset, we adopt the standard output format of extractive MRC tasks,
%(e.g., \squad, \newsqa), 
where a system is required to point to a span within a given paragraph $p$ as its prediction.
However, we define two difficulty levels in \imrc, which are based on different action spaces and dynamics during the interactive information gathering phase.

\textbf{Easy Mode:} 
At a step $t$, an agent can issue one of the following four actions to interact with the (partially observable) paragraph $p$, where $p$ consists of $n$ sentences. Assume the agent's observation $o_t$ corresponds to sentence $s_k$, where $1 \leq k \leq n$.

\begin{itemize}
    \item \cmd{previous}: jump to
        $ %\small
          \begin{cases}
              s_n & \text{if $k = 1$,}\\
              s_{k-1} & \text{otherwise;}
          \end{cases}
        $
    \item \cmd{next}: jump to
        $ %\small
          \begin{cases}
              s_1 & \text{if $k = n$,}\\
              s_{k+1} & \text{otherwise;}
          \end{cases}
        $
    \item \cmd{\ctrlf} \query: jump to the sentence that contains the next occurrence of \query;
    \item \cmd{stop}: terminate information gathering phase and ready to answer question.
\end{itemize}

\textbf{Hard Mode:}
Only the \cmd{\ctrlf} and \cmd{stop} commands are available (i.e., an agent is forced to generate \query to navigate the partially observable paragraph $p$).

\subsection{\query Types}
\label{section:query}

Given an objective (e.g., a question to answer), humans search by using both extractive and abstractive queries. 
For instance, when searching information about the actor ``Dwayne Johnson'', one may either type his name or ``The Rock'' in a search engine. 
We believe abstractive query searching requires a deeper understanding of the question, and some background knowledge (one cannot refer to ``Dwayne Johnson'' as the ``The Rock'' if they know nothing about his wrestling career).

Inspired by this observation, we study the following three settings, where in each, the \query is generated from different sources:
\begin{enumerate}
    \item One token from the question: extractive \query generation with a relatively small action space.
    \item One token from the union of the question and the current observation: still extractive \query generation, although in an intermediate level where the action space is larger.
    \item One token from the dataset vocabulary: abstractive \query generation where the action space is huge (see Table~\ref{tab:dataset_stats} for statistics of \isquad and \inewsqa). 
\end{enumerate}

% \textbf{1.} One token from the question: extractive \query generation with a relatively small action space.\\
% \textbf{2.} One token from the union of the question and the current observation: still extractive \query generation, although in an intermediate level where the action space is larger. \\
% \textbf{3.} One token from the dataset vocabulary: abstractive \query generation where the action space is huge (see Table~\ref{tab:dataset_stats} for statistics of \isquad and \inewsqa). 

% \begin{enumerate}
%     \item One token from the question: extractive \query generation with relatively small action space.
%     \item One token from the union of the question and the current observation: still extractive \query generation, although in an intermediate level where the action space is larger. 
%     \item One token from the dataset vocabulary: abstractive \query generation where the action space is huge (see Table~\ref{tab:dataset_stats} for statistics of \isquad and \inewsqa). 
% \end{enumerate}

\subsection{Evaluation Metric}
\label{section:metric}
Since \imrc involves both MRC and RL, we adopt evaluation metrics from both settings.
First, as a question answering task, we use \fone score to compare predicted answers against ground-truth, as in previous work. 
When there exist multiple ground-truth answers, we report the max \fone score.

Second, mastering multiple games remains quite challenging for RL agents.
Therefore, we evaluate an agent's performance during both its training and testing phases.
Specifically, we report training curves and test results based on the best validation \fone scores.
% During training, we report training curves averaged over 3 random seeds.
%During test, we follow common practice in supervised learning tasks where we report the agent's test performance corresponding to its best validation performance \footnote{Since \squad's test set is hidden.
% We split its training set to get a validation set and use the original development set for testing. We split on the article level to prevent overlap between the training and validation paragraphs. This yields 5,158 validation points.}. 

\begin{table}[t!]
    \centering
    \scriptsize
    \begin{tabular}{c|c|c}
        \toprule
        Dataset &                       \isquad &        \inewsqa  \\
        \midrule
        \#Training Games &              82,441 &        92,550 \\
        \midrule
        Vocabulary Size &               109,689 &       200,000 \\
        \midrule
        Avg. \#Sentence / Document &    5.1 &           29.5 \\
        \midrule
        Avg. Sentence Length &         26.1 &           22.2 \\
        \midrule
        Avg. Question Length &         11.3 &           7.6 \\
        \bottomrule
    \end{tabular}
    \caption{Statistics of \isquad and \inewsqa.}
    \label{tab:dataset_stats}
\end{table}

\section{Baseline Agent}
\label{section:baseline}

As a baseline agent, we adopt \qadqn \citep{yuan2019qait}, we modify it to enable extractive \query generation and question answering. 

As illustrated in Figure~\ref{fig:model}, the baseline agent consists of three components: an encoder, an action generator, and a question answerer. 
More precisely, at a step $t$ during the information gathering phase, the encoder reads observation string $o_t$ and question string $q$ to generate the attention aggregated hidden representations $M_t$. 
Using $M_t$, the action generator outputs commands (depending on the mode, as defined in \S~\ref{section:easy_hard_settings}) to interact with \imrc.
The information-gathering phase terminates whenever the generated command is \cmd{stop} or the agent has used up its move budget.
The question answerer takes the hidden representation at the terminating step to generate head and tail pointers as its answer prediction.

\subsection{Model Structure}

In this section, we only describe the \textbf{difference} between the model our baseline agent uses and the original \qadqn. We refer readers to \citep{yuan2019qait} for detailed information.

In the following subsections, we use ``game step $t$'' to denote the $t$th round of interaction between an agent with the \imrc environment.

\subsubsection{Action Generator}

Let $M_t \in \mathbb{R}^{L \times H}$ denote the output of the encoder, where $L$ is the length of observation string and $H$ is hidden size of the encoder representations.

The action generator takes $M_t$ as input and generates rankings for all possible actions. 
As described in the previous section, a \cmd{\ctrlf} command is composed of two tokens (the token ``\cmd{\ctrlf}'' and the \query token).
Therefore, the action generator consists of three multilayer perceptrons (MLPs): 
%$L_{shared}$ is shared; $L_{action}$ has size of number of all possible actions; $L_{ctrlf}$ has size of all possible tokens that can be used as \cmd{query}:
\begin{equation}
\label{eqn:qvalue}
% \small
\begin{aligned}
R_t &= \relu(\mlp_{\text{shared}}(\text{mean}(M_t))), \\
Q_{t, \text{action}} &= \mlp_{\text{action}}(R_t) \cdot M_\text{mode}, \\
Q_{t, \text{query}} &= \mlp_{\text{query}}(R_t) \cdot M_\text{type}. \\
\end{aligned}
\end{equation}
In which, $Q_{t, \text{action}}$ and $Q_{t, \text{query}}$ are Q-values of action token and \query token (when action token is ``\cmd{\ctrlf}''), respectively. 
$M_\text{mode}$ is a mask, which masks the \cmd{previous} and \cmd{next} tokens in hard mode;
$M_\text{type}$ is another mask which depends on the current \query type (e.g., when \query is extracted from the question $q$, all tokens absent from $q$ are masked out).
Probability distributions of tokens are further computed by applying softmax on $Q_{t, \text{action}}$ and $Q_{t, \text{query}}$, respectively.
% \begin{equation}
% \small
% \begin{aligned}
% p_{t, \text{action}} = &~\smx(Q_{t, \text{action}}), \\
% p_{t, \text{query}} = &~\smx(Q_{t, \text{query}}). \\
% \end{aligned}
% \end{equation}

\subsubsection{Question Answerer}
Following QANet \citep{yu18qanet}, we append two extra stacks of transformer blocks on top of the encoder to compute head and tail positions:
\begin{equation}
\label{eqn:qa_mlp01}
% \small
\begin{aligned}
h_{\text{head}} &= \relu(\mlp_0(\lbrack M_t; M_{\text{head}}\rbrack)), \\
h_{\text{tail}} &= \relu(\mlp_1(\lbrack M_t; M_{\text{tail}}\rbrack)). \\
\end{aligned}
\end{equation}
In which, $[\cdot;\cdot]$ denotes vector concatenation, $M_{\text{head}} \in \mathbb{R}^{L \times H}$ and $M_{\text{tail}} \in \mathbb{R}^{L \times H}$ are the outputs of the two extra transformer stacks.

Similarly, probability distributions of head and tail pointers over observation string $o_t$ can be computed by:
\begin{equation}
\label{eqn:qa_mlp23}
% \small
\begin{aligned}
p_{\text{head}} &= \smx(\mlp_2(h_{\text{head}})), \\
p_{\text{tail}} &= \smx(\mlp_3(h_{\text{tail}})). \\
\end{aligned}
\end{equation}

\subsection{Memory and Reward Shaping}

\subsubsection{Memory}
In \imrc tasks, some questions may not be easily answerable by observing a single sentence.
To overcome this limitation, we provide an explicit memory mechanism to our baseline agent to serve as an inductive bias. 
Specifically, we use a queue to store strings that have been observed recently.
The queue has a limited number of slots (we use queues of size [1, 3, 5] in this work).
This prevents the agent from issuing \cmd{next} commands until the environment is observed fully in memory, in which case our task degenerates to the standard MRC setting.
We reset the memory slots episodically.

\subsubsection{Reward Shaping}
\label{section:reward_shaping}
Because the question answerer in our agent is a pointing model, its performance relies heavily on whether the agent can find and stop at the sentence that contains the answer.
In the same spirit as \citep{yuan2019qait}, we also design a heuristic reward to guide agents to learn this behavior.

In particular, we assign a reward if the agent halts at game step $k$ and the answer is a sub-string of $o_k$ (if larger memory slots are used, we assign this reward if the answer is a sub-string of the memory at game step $k$). 
We denote this reward as the \textbf{sufficient information reward}, since, if an agent sees the answer, it should have a good chance of having gathered sufficient information for the question (although this is not guaranteed). 

Note this sufficient information reward is part of the design of the baseline agent, whereas the question answering score is the only metric used to evaluate an agent's performance on the \imrc task.

\subsection{Training Strategy}
Since \imrc games are interactive environments and we have formulated the tasks as POMDPs (in \S~\ref{section:pomdp}), it is natural to use RL algorithms to train the information gathering components of our agent. 
In this work, we study the performance of two widely used RL algorithms, one based on Q-Learning (DQN) and the other on Policy Gradients (A2C).
When an agent has reached a sentence that contains sufficient information to answer the question, the task becomes a standard extractive QA setting, where an agent learns to point to a span from its observation.
When this condition is met, it is also natural to adopt standard supervised learning methods to train the question answering component of our agent.

In this section, we describe the 3 training strategies mentioned above. 
We provide implementation details in Appendix~\ref{appd:implementation_details}.

\begin{figure}[t!]
    \centering
    \includegraphics[width=0.5\textwidth]{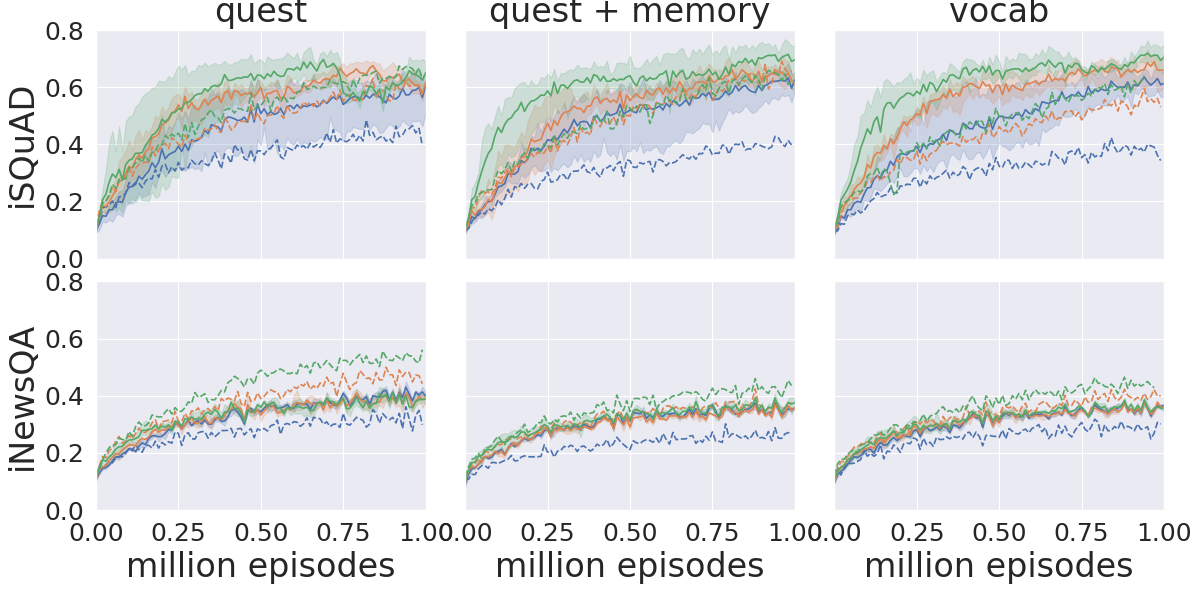}
    \caption{Training \fone scores in \textbf{easy mode} with different \query types and memory sizes. Solid line: DQN, dashed line: A2C; number of memory slots:  \textcolor{blue1}{1}, \textcolor{orange}{3}, \textcolor{green1}{5}.}
    \label{fig:squad_newsqa_train}
\end{figure}

\subsubsection{Advantage Actor-Critic (A2C)}
Advantage actor-critic (A2C) was first proposed by \citet{mnih16a3c}. 
Compared to policy gradient computation in REINFORCE \citep{williams1992reinforce},
\begin{equation}
% \small
    \nabla_\theta J(\theta) = \mathbb{E}_\pi [\sum_{t=1}^{T}\nabla_\theta \log \pi_\theta(a_t| s_t) G_t],
    \label{eq:pg_reinforce}
\end{equation}
where the policy gradient $\nabla_\theta J(\theta)$ is updated by measuring the discounted future reward $G_t$ from real sample trajectories, A2C utilizes the lower variance advantage function $A(s_t, a_t) = Q(s_t, a_t) - V(s_t)$ in place of $G_t$. 
The advantage $A(s_t, a_t)$ of taking action $a_t$ at state $s_t$ is defined as the value $Q(s_t, a_t)$ of taking $a_t$ minus the average value $V(s_t)$ of all possible actions in state $s_t$.
% In which, $Q(s_t, a_t) = \mathbb{E}[r_{t+1} + \gamma V(s_{t+1})]$.

In the agent, a critic updates the state-value function $V(s)$, whereas an actor updates the policy parameter $\theta$ for $\pi_\theta(a|s)$, in the direction suggested by the critic.
Following common practice, we share parameters between actor and critic networks. 
Specifically, all parameters other than $\mlp_{\text{action}}$ and $\mlp_{\text{query}}$ (both defined in Eqn.~\ref{eqn:qvalue}) are shared between actor and critic.
% Specifically, all parameters up to $\mlp_{\text{shared}}$ as described in Eqn.~\ref{eqn:qvalue} are shared and the actor and critic both have their own $\mlp_{\text{action}}$ and $\mlp_{\text{query}}$.

\subsubsection{Deep Q-Networks (DQN)}

In Q-Learning \citep{watkins1992qlearning, mnih2015dqn}, given an interactive environment, an agent takes an action $a_t$ in state $s_t$ by consulting a state-action value estimator $Q(s, a)$; this value estimator estimates the action's expected long-term reward. 
Q-Learning helps the agent to learn an optimal value estimator. 
An agent starts from performing randomly and gradually updates its value estimator by interacting with the environment and propagating reward information. 
In our case, the estimated Q-value at game step $t$ is simply the sum of Q-values of the action token and \query token as introduced in Eqn.~\ref{eqn:qvalue}: %$Q_{t} = Q_{t, \text{action}} + Q_{t, \text{query}}$.
\begin{equation}
Q_{t} = Q_{t, \text{action}} + Q_{t, \text{query}}. \\
\end{equation}

\begin{figure}[t!]
    \centering
    \includegraphics[width=0.5\textwidth]{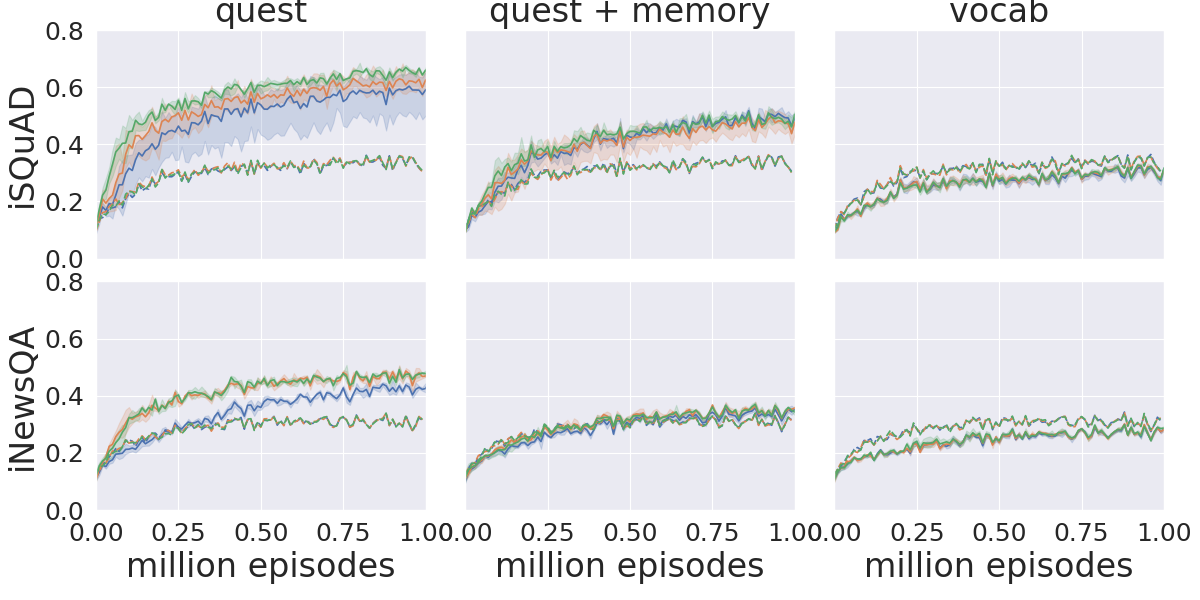}
    \caption{Training \fone scores in \textbf{hard mode} with different \query types and memory sizes. Solid line: DQN, dashed line: A2C; number of memory slots:  \textcolor{blue1}{1}, \textcolor{orange}{3}, \textcolor{green1}{5}.}
    \label{fig:squad_newsqa_train_ctrlf}
\end{figure}

In this work, we adopt the Rainbow algorithm \citep{hessel17rainbow}, which is a deep Q-network boosted by several extensions such as a prioritized replay buffer \citep{schaul2016replay}. 
Rainbow exhibits state-of-the-art performance on several RL benchmark tasks (e.g., Atari games).

\subsubsection{Negative Log-likelihood (NLL)}
During information gathering phase, we use another replay buffer to store question answering transitions (observation string when interaction stops, question string, ground-truth answer) whenever the terminal observation string contains the ground-truth answer.
We randomly sample mini-batches of such transitions to train the question answerer to minimize the negative log-likelihood loss.

\section{Experimental Results}
\label{section:exp}

\begin{table*}[t!]
    \centering
    \scriptsize
    \begin{tabular}{c|c|c|c|c||c|c|c}
        \toprule
        \multicolumn{2}{c|}{Dataset} & \multicolumn{3}{c||}{\isquad} & \multicolumn{3}{c}{\inewsqa}\\
        \midrule % 
        \multicolumn{8}{c}{Easy Mode} \\
        \midrule % 
        \query                                  &       Agent    & Mem=1                        &    =3 &    =5 & Mem=1 &    =3 &    =5 \\
        \midrule % 
        Question                                &       \aac     & 0.245 \color{blue}(0.493)    & 0.357 \color{blue}(0.480) & 0.386 \color{blue}(0.478) & 0.210 \color{blue}(0.554) & 0.316 \color{blue}(0.532) & 0.333 \color{blue}(0.490)  \\
                                                &       \dqn     & 0.575 \color{blue}(0.770)    & 0.637 \color{blue}(0.738) & \textbf{0.666} \color{blue}(0.716) & 0.330 \color{blue}(0.708) & 0.326 \color{blue}(0.619) & \textbf{0.360} \color{blue}(0.620)  \\
        \midrule % 
        Question+Memory                         &       \aac     & 0.221 \color{blue}(0.479)    & 0.484 \color{blue}(0.590) & 0.409 \color{blue}(0.492) & 0.199 \color{blue}(0.595) & 0.233 \color{blue}(0.448) & 0.253 \color{blue}(0.459)  \\
                                                &       \dqn     & 0.579 \color{blue}(0.784)    & 0.651 \color{blue}(0.734) & 0.656 \color{blue}(0.706) & 0.336 \color{blue}(0.715) & 0.334 \color{blue}(0.626) & 0.347 \color{blue}(0.596)  \\
        \midrule % 
        Vocabulary                              &       \aac     & 0.223 \color{blue}(0.486)    & 0.314 \color{blue}(0.448) & 0.309 \color{blue}(0.391) & 0.192 \color{blue}(0.551) & 0.224 \color{blue}(0.440) & 0.224 \color{blue}(0.403)  \\
                                                &       \dqn     & 0.583 \color{blue}(0.774)    & 0.624 \color{blue}(0.738) & 0.661 \color{blue}(0.731) & 0.326 \color{blue}(0.715) & 0.323 \color{blue}(0.590) & 0.316 \color{blue}(0.593)  \\
        \midrule % 
        \multicolumn{8}{c}{Hard Mode}  \\
        \midrule % 
        Question                                &       \aac     & 0.147 \color{blue}(0.404)    & 0.162 \color{blue}(0.446) & 0.158 \color{blue}(0.435) & 0.166 \color{blue}(0.529) & 0.160 \color{blue}(0.508) & 0.164 \color{blue}(0.520)  \\
                                                &       \dqn     & 0.524 \color{blue}(0.766)    & 0.524 \color{blue}(0.740) & \textbf{0.551} \color{blue}(0.739) & 0.352 \color{blue}(0.716) & \textbf{0.367} \color{blue}(0.632) & 0.353 \color{blue}(0.613)  \\
        \midrule % 
        Question+Memory                         &       \aac     & 0.160 \color{blue}(0.441)    & 0.150 \color{blue}(0.413) & 0.156 \color{blue}(0.429) & 0.163 \color{blue}(0.520) & 0.160 \color{blue}(0.508) & 0.164 \color{blue}(0.520)  \\
                                                &       \dqn     & 0.357 \color{blue}(0.749)    & 0.362 \color{blue}(0.729) & 0.364 \color{blue}(0.733) & 0.260 \color{blue}(0.692) & 0.264 \color{blue}(0.645) & 0.269 \color{blue}(0.620)  \\
        \midrule % 
        Vocabulary                              &       \aac     & 0.161 \color{blue}(0.444)    & 0.163 \color{blue}(0.448) & 0.160 \color{blue}(0.441) & 0.160 \color{blue}(0.510) & 0.167 \color{blue}(0.532) & 0.162 \color{blue}(0.516)  \\
                                                &       \dqn     & 0.264 \color{blue}(0.728)    & 0.261 \color{blue}(0.719) & 0.218 \color{blue}(0.713) & 0.326 \color{blue}(0.694) & 0.214 \color{blue}(0.680) & 0.214 \color{blue}(0.680)  \\
        \bottomrule
    \end{tabular}
    \caption{Test \fone scores in \textbf{black} and \foneinfo scores (i.e., an agent's \fone score iff sufficient information is in its observation when it terminates information gathering phase) in \textcolor{blue}{\textbf{blue}}.}
    \label{tab:result_table}
\end{table*}

In this study, we focus on four main aspects:
\begin{enumerate}
    \item difficulty levels (easy $|$ hard mode);
    \item strategies for generating \query (from question $|$ question and observation $|$ vocabulary);
    \item sizes of the memory queue (1 $|$ 3 $|$ 5);
    \item RL algorithms for the information gathering phase (A2C $|$ DQN)
\end{enumerate}

% \textbf{1.} difficulty levels (easy $|$ hard mode);\\
% \textbf{2.} strategies for generating \query (from question $|$ question and observation $|$ vocabulary);\\
% \textbf{3.} sizes of the memory queue (1 $|$ 3 $|$ 5);\\
% \textbf{4.} RL algorithms for the information gathering phase (A2C $|$ DQN).
% \begin{enumerate}
%     \item different difficulty levels (easy / hard mode);
%     \item different strategies for generating \query (from question / question and observation / vocabulary );
%     \item different sizes of memory slots (1 / 3 / 5);
%     \item different RL algorithms (A2C / DQN).
% \end{enumerate}

Regarding the four aspects, we report the baseline agent's training performance followed by its generalization performance on test data.
We use \dqn and \aac to refer to our baseline agent trained with DQN and A2C, respectively.

We set the maximum number of episodes (data points) to be 1 million, this is approximately 10 epochs in supervised learning tasks given the size of datasets.
The agent may further improve after 1 million episodes, however we believe some meaningful and interesting trends can already be observed from the results. 
Besides, we hope to keep the wall clock time of the task reasonable\footnote{Basic experiment setting (e.g., \query from question, single slot memory) take about a day on a single NVIDIA P100 GPU.} to encourage the community to work on this direction.

\subsection{Mastering Training Games}

It remains difficult for RL agents to master multiple games at the same time. 
In our case, each document-question pair can be considered a unique ``game,'' and there are hundreds of thousands of them. 
Therefore, as it is common practice in the RL literature, we study an agent's training curves.

Figure~\ref{fig:squad_newsqa_train} and Figure~\ref{fig:squad_newsqa_train_ctrlf} show the agent's training performance (in terms of \fone score) in easy and hard mode, respectively.
Due to the space limitations, we select several representative settings to discuss in this section.
We provide the agent's training and validation curves for all experiments, and its sufficient information rewards (as defined in \S~\ref{section:reward_shaping}) in Appendix~\ref{appd:more_results}.
% We also provide the agent's sufficient information rewards (as defined in \S~\ref{section:reward_shaping}) in Appendix~\ref{appd:more_results}.
% Note that all training curves are averaged over 3 runs with different random seeds and all evaluation curves show the one run with max validation performance among the three. 

% easy vs hard
It is clear that our agent performs better on easy mode consistently across both datasets and all training strategies.
This may due to the fact that the \cmd{previous} and \cmd{next} commands provide the agent an inefficient but guaranteed way to %visit any sentence in all games. 
stumble on the sought-after sentence no matter the game. 
The \cmd{\ctrlf} command matches human behavior more closely, but it is arguably more challenging (and interesting) for an RL agent to learn this behavior.
RL agents may require extra effort and time to reach a desired state since they rely heavily on random exploration, and the \cmd{\ctrlf} command leads to much larger action space to explore compared to commands such as \cmd{next}.

% question / question and observation / vocabulary
Related to action space size, we observe that the agent performs best when pointing to the \query tokens from the question, whereas it performs worst when generating \query tokens from the entire vocabulary.
This is particularly clear in hard mode, where agents are forced to use the \cmd{\ctrlf} command.
As shown in Table~\ref{tab:dataset_stats}, both datasets have a vocabulary size of more than 100k, whereas the average length of questions is around 10 tokens.
This indicates the action space for generating \query from entire vocabulary is much larger. %than the other two.
% As also shown in Table~\ref{tab:dataset_stats}, the average length of questions is around 10 tokens.
This again suggests that for moving toward a more realistic problem setting where action spaces are huge, methods with better sample efficiency are needed.

% different sizes of memory slots (1 / 3 / 5)
Experiments show that a larger memory queue almost always helps. 
Intuitively, with a memory mechanism (either explicit as in this work, or implicit as with a recurrent network aggregating representations over game steps), an agent renders the environment closer to fully observed by exploring and storing observations. 
Presumably, a larger memory could further improve an agent's performance; considering the average number of sentences in each \isquad game is 5, a memory with more than 5 slots defeats the purpose of our study of partially observable text environments.

% A2C vs DQN
We observe that \dqn generally performs better on \isquad whereas \aac sometimes works better on the harder \inewsqa task.
However, we observe huge gap between them on generalization performance, which we discuss in a later subsection.

% iSquad vs iNewsQA
Not surprisingly, our agent performs better in general on \isquad than on \inewsqa.
As shown in Table~\ref{tab:dataset_stats}, the average number of sentences per document in \inewsqa is about 6 times more than in \isquad. 
This is analogous to partially observable games with larger maps in the RL literature.
We believe a better exploration (in our case, jumping) strategy that can decide where to explore next conditioned on what has already been seen may help agents to master such harder games.

\subsection{Generalizing to Test Set}

To study an agent's ability to generalize, we select the best performing checkpoint in each experimental setting on the validation set and report their test performance, as shown in Table~\ref{tab:result_table}. 
In addition, to support our claim that the more challenging part of \imrc tasks is information gathering rather than answering questions given sufficient information, we report the agents' \fone scores when they have reached the piece of text that contains the answer, which we denote as \foneinfo.

%  different difficulty levels (easy / hard mode)
%  different strategies for generating \query (from question / question and observation / vocabulary )
From Table~\ref{tab:result_table} (and validation curves provided in Appendix~\ref{appd:more_results}) we observe trends that match with training curves.
Due to the different sizes of action space, the baseline agents consistently performs better on the easy mode. 
For the same reason, the agent learns more efficiently when the \query token is extracted from the question.
The best \fone score on hard mode is comparable to and even slightly higher than in easy mode on \inewsqa, which suggests our baseline agent learns some relatively general trajectories in solving training games that generalize to unseen games.

%  different sizes of memory slots (1 / 3 / 5)
It is also clear that during evaluation, a memory that stores experienced observations helps, since the agent almost always performs better with a memory size of 3 or 5 (when memory size is 1, each new observation overwrites the memory).

%  different RL algorithms during information gathering phase (A2C / DQN)
While performing comparably with \dqn during training, the agent trained with \aac generalizes noticeably worse.
We suspect this is caused by the fundamental difference between the ways \dqn and \aac explore during training.
Specifically, \dqn relies on either $\epsilon$-greedy or Noisy Net \citep{fortunato17noisynet}, both of which explicitly force an agent to experience different actions during training. In \aac, exploration is performed implicitly by sampling from a probability distribution over the action space; although entropy regularization is applied, good exploration is still not guaranteed (if there are peaks in the probability distribution).
This again suggests the importance of a good exploration strategy in the \imrc tasks, as in all RL tasks.

Finally, we observe \foneinfo scores are consistently higher than the overall \fone scores, and they have less variance across different settings.
This supports our hypothesis that information gathering plays an important role in solving \imrc tasks, whereas question answering given necessary information is relatively straightforward. 
% This also suggests that an interactive agent that can better navigate to important sentences is very likely to achieve better performance on \imrc tasks.

\section{Discussion and Future Work}
\label{section:discuss}

In this work, we propose and explore the direction of converting MRC datasets into interactive, partially observable environments. 
We believe information-seeking behavior is desirable for neural MRC systems when knowledge sources are partially observable and/or too large to encode in their entirety, 
% when searching for information on the internet, 
where knowledge is by design easily accessible to humans through interaction. 
Our idea for reformulating existing MRC datasets as partially observable and interactive environments is straightforward and general. 
It is complementary to existing MRC dataset and models, meaning almost all MRC datasets can be used to study interactive, information-seeking behavior through similar modifications.
We hypothesize that such behavior can, in turn, help in solving real-world MRC problems involving search.
As a concrete example, in real world environments such as the Internet, different pieces of knowledge are interconnected by hyperlinks. 
We could equip the agent with an action to ``click'' a hyperlink, which returns another webpage as new observations, thus allowing it to navigate through a large number of web information to answer difficult questions. 

%In such environments, an interactive information-seeking system can leverage the above-mentioned inductive bias and sequentially making decisions about what to read next (e.g., which hyperlink to jump to), and eventually reach necessary information.

\imrc is difficult and cannot yet be solved, however it clearly matches a human's information-seeking behavior compared to most static and fully-observable laboratory MRC benchmarks.
%Despite being restricted, our proposed task presents major challenges to existing techniques.
It lies at the intersection of NLP and RL, which is arguably less studied in existing literature.
%We hope to encourage researchers from both NLP and RL communities to work toward solving this task.
For our baseline, we adopted off-the-shelf, top-performing MRC and RL methods, and applied a memory mechanism which serves as an inductive bias.
Despite being necessary, our preliminary experiments do not seem sufficient.
We encourage work on this task to determine what inductive biases, architectural components, or pretraining recipes are necessary or sufficient for MRC based on information-seeking.

% Either component can be replaced straightforwardly with other methods (e.g., to utilize a large-scale pretrained language model).

Our proposed setup presently uses only a single word as \query in the \cmd{\ctrlf} command in an abstractive manner. 
However, a host of other options could be considered in future work. 
For example, a multi-word \query with fuzzy matching is more realistic.
It would also be interesting for an agent to generate a vector representation of the \query in some latent space and modify it during the dynamic reasoning process.
This could further be used to retrieve different contents by comparing with pre-computed document representations (e.g., in an open-domain QA dataset),
% This could then be compared with pre-computed document representations (e.g., in an open domain QA dataset) to determine what text to observe next, 
with such behavior tantamount to learning to do IR.
This extends traditional query reformulation for open-domain QA by allowing to drastically change the queries without strictly keeping the semantic meaning of the original queries.

\section*{Acknowledgments}
The authors thank Mehdi Fatemi, Peter Potash, Matthew Hausknecht, and Philip Bachman for insightful ideas and discussions.
We also thank the anonymous ACL reviewers for their helpful feedback and suggestions.

\bibliography{gamify_bib.bib}

\begin{thebibliography}{31}
\expandafter\ifx\csname natexlab\endcsname\relax\def\natexlab#1{#1}\fi

\bibitem[{Bachman et~al.(2016)Bachman, Sordoni, and Trischler}]{Bachman2016}
Philip Bachman, Alessandro Sordoni, and Adam Trischler. 2016.
\newblock Towards information-seeking agents.
\newblock \emph{arXiv preprint arXiv:1612.02605}.

\bibitem[{Choi et~al.(2017)Choi, Hewlett, Uszkoreit, Polosukhin, Lacoste, and
  Berant}]{Choi2017}
Eunsol Choi, Daniel Hewlett, Jakob Uszkoreit, Illia Polosukhin, Alexandre
  Lacoste, and Jonathan Berant. 2017.
\newblock Coarse-to-fine question answering for long documents.
\newblock In \emph{Proceedings of the 55th Annual Meeting of the Association
  for Computational Linguistics (Volume 1: Long Papers)}, volume~1, pages
  209--220.

\bibitem[{Fortunato et~al.(2017)Fortunato, Azar, Piot, Menick, Osband, Graves,
  Mnih, Munos, Hassabis, Pietquin, Blundell, and Legg}]{fortunato17noisynet}
Meire Fortunato, Mohammad~Gheshlaghi Azar, Bilal Piot, Jacob Menick, Ian
  Osband, Alex Graves, Vlad Mnih, R{\'{e}}mi Munos, Demis Hassabis, Olivier
  Pietquin, Charles Blundell, and Shane Legg. 2017.
\newblock \href {http://arxiv.org/abs/1706.10295} {Noisy networks for
  exploration}.
\newblock \emph{CoRR}, abs/1706.10295.

\bibitem[{Geva and Berant(2018)}]{geva2018learning}
Mor Geva and Jonathan Berant. 2018.
\newblock Learning to search in long documents using document structure.
\newblock \emph{arXiv preprint arXiv:1806.03529}.

\bibitem[{Han et~al.(2019)Han, Kang, Jung, and Hwang}]{Han2019}
Moonsu Han, Minki Kang, Hyunwoo Jung, and Sung~Ju Hwang. 2019.
\newblock Episodic memory reader: Learning what to remember for question
  answering from streaming data.
\newblock \emph{arXiv preprint arXiv:1903.06164}.

\bibitem[{Hansen et~al.(2019)Hansen, Hansen, Alstrup, Simonsen, and
  Lioma}]{Hansen2019}
Christian Hansen, Casper Hansen, Stephen Alstrup, Jakob~Grue Simonsen, and
  Christina Lioma. 2019.
\newblock Neural speed reading with structural-jump-lstm.
\newblock \emph{arXiv preprint arXiv:1904.00761}.

\bibitem[{Hessel et~al.(2017)Hessel, Modayil, van Hasselt, Schaul, Ostrovski,
  Dabney, Horgan, Piot, Azar, and Silver}]{hessel17rainbow}
Matteo Hessel, Joseph Modayil, Hado van Hasselt, Tom Schaul, Georg Ostrovski,
  Will Dabney, Daniel Horgan, Bilal Piot, Mohammad~Gheshlaghi Azar, and David
  Silver. 2017.
\newblock \href {http://arxiv.org/abs/1710.02298} {Rainbow: Combining
  improvements in deep reinforcement learning}.
\newblock \emph{CoRR}, abs/1710.02298.

\bibitem[{Kaelbling et~al.(1998)Kaelbling, Littman, and
  Cassandra}]{kaelbling1998planning}
Leslie~Pack Kaelbling, Michael~L Littman, and Anthony~R Cassandra. 1998.
\newblock Planning and acting in partially observable stochastic domains.
\newblock \emph{Artificial intelligence}, 101(1-2):99--134.

\bibitem[{Kingma and Ba(2014)}]{kingma14adam}
Diederik~P Kingma and Jimmy Ba. 2014.
\newblock Adam: A method for stochastic optimization.
\newblock \emph{arXiv preprint arXiv:1412.6980}.

\bibitem[{Mikolov et~al.(2018)Mikolov, Grave, Bojanowski, Puhrsch, and
  Joulin}]{mikolov18fasttext}
Tomas Mikolov, Edouard Grave, Piotr Bojanowski, Christian Puhrsch, and Armand
  Joulin. 2018.
\newblock Advances in pre-training distributed word representations.
\newblock In \emph{Proceedings of the International Conference on Language
  Resources and Evaluation (LREC 2018)}.

\bibitem[{Mnih et~al.(2016)Mnih, Badia, Mirza, Graves, Lillicrap, Harley,
  Silver, and Kavukcuoglu}]{mnih16a3c}
Volodymyr Mnih, Adri{\`{a}}~Puigdom{\`{e}}nech Badia, Mehdi Mirza, Alex Graves,
  Timothy~P. Lillicrap, Tim Harley, David Silver, and Koray Kavukcuoglu. 2016.
\newblock \href {http://arxiv.org/abs/1602.01783} {Asynchronous methods for
  deep reinforcement learning}.
\newblock \emph{CoRR}, abs/1602.01783.

\bibitem[{Mnih et~al.(2015)Mnih, Kavukcuoglu, Silver, Rusu, Veness, Bellemare,
  Graves, Riedmiller, Fidjeland, Ostrovski et~al.}]{mnih2015dqn}
Volodymyr Mnih, Koray Kavukcuoglu, David Silver, Andrei~A Rusu, Joel Veness,
  Marc~G Bellemare, Alex Graves, Martin Riedmiller, Andreas~K Fidjeland, Georg
  Ostrovski, et~al. 2015.
\newblock Human-level control through deep reinforcement learning.
\newblock \emph{Nature}, 518(7540):529--533.

\bibitem[{Narasimhan et~al.(2016)Narasimhan, Yala, and
  Barzilay}]{narasimhan16acquiring}
Karthik Narasimhan, Adam Yala, and Regina Barzilay. 2016.
\newblock \href {http://arxiv.org/abs/1603.07954} {Improving information
  extraction by acquiring external evidence with reinforcement learning}.
\newblock \emph{CoRR}, abs/1603.07954.

\bibitem[{Nguyen et~al.(2016)Nguyen, Rosenberg, Song, Gao, Tiwary, Majumder,
  and Deng}]{nguyen16msmarco}
Tri Nguyen, Mir Rosenberg, Xia Song, Jianfeng Gao, Saurabh Tiwary, Rangan
  Majumder, and Li~Deng. 2016.
\newblock \href {http://arxiv.org/abs/1611.09268} {{MS} {MARCO:} {A} human
  generated machine reading comprehension dataset}.
\newblock \emph{CoRR}, abs/1611.09268.

\bibitem[{Nogueira and Cho(2016)}]{nogueira16web}
Rodrigo Nogueira and Kyunghyun Cho. 2016.
\newblock \href {http://arxiv.org/abs/1602.02261} {Webnav: {A} new large-scale
  task for natural language based sequential decision making}.
\newblock \emph{CoRR}, abs/1602.02261.

\bibitem[{Nogueira and Cho(2017)}]{nogueira17queryreform}
Rodrigo Nogueira and Kyunghyun Cho. 2017.
\newblock \href {http://arxiv.org/abs/1704.04572} {Task-oriented query
  reformulation with reinforcement learning}.
\newblock \emph{CoRR}, abs/1704.04572.

\bibitem[{Qi et~al.(2019)Qi, Lin, Mehr, Wang, and Manning}]{qi2019answering}
Peng Qi, Xiaowen Lin, Leo Mehr, Zijian Wang, and Christopher~D. Manning. 2019.
\newblock \href {https://nlp.stanford.edu/pubs/qi2019answering.pdf} {Answering
  complex open-domain questions through iterative query generation}.
\newblock In \emph{2019 Conference on Empirical Methods in Natural Language
  Processing and 9th International Joint Conference on Natural Language
  Processing ({EMNLP-IJCNLP})}.

\bibitem[{Rajpurkar et~al.(2016)Rajpurkar, Zhang, Lopyrev, and
  Liang}]{rajpurkar16squad}
Pranav Rajpurkar, Jian Zhang, Konstantin Lopyrev, and Percy Liang. 2016.
\newblock \href {http://arxiv.org/abs/1606.05250} {Squad: 100, 000+ questions
  for machine comprehension of text}.
\newblock \emph{CoRR}, abs/1606.05250.

\bibitem[{Reddy et~al.(2018)Reddy, Chen, and Manning}]{reddy18coqa}
Siva Reddy, Danqi Chen, and Christopher~D. Manning. 2018.
\newblock \href {http://arxiv.org/abs/1808.07042} {Coqa: {A} conversational
  question answering challenge}.
\newblock \emph{CoRR}, abs/1808.07042.

\bibitem[{Schaul et~al.(2016)Schaul, Quan, Antonoglou, and
  Silver}]{schaul2016replay}
Tom Schaul, John Quan, Ioannis Antonoglou, and David Silver. 2016.
\newblock Prioritized experience replay.
\newblock In \emph{International Conference on Learning Representations},
  Puerto Rico.

\bibitem[{Seo et~al.(2017)Seo, Min, Farhadi, and Hajishirzi}]{Seo2017}
Minjoon Seo, Sewon Min, Ali Farhadi, and Hannaneh Hajishirzi. 2017.
\newblock Neural speed reading via skim-rnn.
\newblock \emph{arXiv preprint arXiv:1711.02085}.

\bibitem[{Sha et~al.(2017)Sha, Qian, and Sui}]{Sha2017}
Lei Sha, Feng Qian, and Zhifang Sui. 2017.
\newblock Will repeated reading benefit natural language understanding?
\newblock In \emph{National CCF Conference on Natural Language Processing and
  Chinese Computing}, pages 366--379. Springer.

\bibitem[{Shen et~al.(2017)Shen, Huang, Gao, and Chen}]{Shen2017}
Yelong Shen, Po-Sen Huang, Jianfeng Gao, and Weizhu Chen. 2017.
\newblock Reasonet: Learning to stop reading in machine comprehension.
\newblock In \emph{Proceedings of the 23rd ACM SIGKDD International Conference
  on Knowledge Discovery and Data Mining}, pages 1047--1055. ACM.

\bibitem[{Sugawara et~al.(2018)Sugawara, Inui, Sekine, and
  Aizawa}]{sugawara18easier}
Saku Sugawara, Kentaro Inui, Satoshi Sekine, and Akiko Aizawa. 2018.
\newblock \href {http://arxiv.org/abs/1808.09384} {What makes reading
  comprehension questions easier?}
\newblock \emph{CoRR}, abs/1808.09384.

\bibitem[{Trischler et~al.(2016)Trischler, Wang, Yuan, Harris, Sordoni,
  Bachman, and Suleman}]{trischler16newsqa}
Adam Trischler, Tong Wang, Xingdi Yuan, Justin Harris, Alessandro Sordoni,
  Philip Bachman, and Kaheer Suleman. 2016.
\newblock \href {http://arxiv.org/abs/1611.09830} {Newsqa: {A} machine
  comprehension dataset}.
\newblock \emph{CoRR}, abs/1611.09830.

\bibitem[{Watkins and Dayan(1992)}]{watkins1992qlearning}
Christopher J. C.~H. Watkins and Peter Dayan. 1992.
\newblock \href {https://doi.org/10.1007/BF00992698} {Q-learning}.
\newblock \emph{Machine Learning}, 8(3):279--292.

\bibitem[{Williams(1992)}]{williams1992reinforce}
Ronald~J Williams. 1992.
\newblock Simple statistical gradient-following algorithms for connectionist
  reinforcement learning.
\newblock \emph{Machine learning}, 8(3-4):229--256.

\bibitem[{Yang et~al.(2018)Yang, Qi, Zhang, Bengio, Cohen, Salakhutdinov, and
  Manning}]{yang18hotpot}
Zhilin Yang, Peng Qi, Saizheng Zhang, Yoshua Bengio, William~W. Cohen, Ruslan
  Salakhutdinov, and Christopher~D. Manning. 2018.
\newblock \href {http://arxiv.org/abs/1809.09600} {Hotpotqa: {A} dataset for
  diverse, explainable multi-hop question answering}.
\newblock \emph{CoRR}, abs/1809.09600.

\bibitem[{Yu et~al.(2018)Yu, Dohan, Luong, Zhao, Chen, Norouzi, and
  Le}]{yu18qanet}
Adams~Wei Yu, David Dohan, Minh{-}Thang Luong, Rui Zhao, Kai Chen, Mohammad
  Norouzi, and Quoc~V. Le. 2018.
\newblock \href {http://arxiv.org/abs/1804.09541} {Qanet: Combining local
  convolution with global self-attention for reading comprehension}.
\newblock \emph{CoRR}, abs/1804.09541.

\bibitem[{Yu et~al.(2017)Yu, Lee, and Le}]{Yu2017}
Adams~Wei Yu, Hongrae Lee, and Quoc~V Le. 2017.
\newblock Learning to skim text.
\newblock \emph{arXiv preprint arXiv:1704.06877}.

\bibitem[{Yuan et~al.(2019)Yuan, C\^ot\'{e}, Fu, Lin, Pal, Bengio, and
  Trischler}]{yuan2019qait}
Xingdi Yuan, Marc-Alexandre C\^ot\'{e}, Jie Fu, Zhouhan Lin, Christopher Pal,
  Yoshua Bengio, and Adam Trischler. 2019.
\newblock Interactive language learning by question answering.

\end{thebibliography}
\bibliographystyle{acl_natbib}

\clearpage
\appendix

\section{Full Results}
\label{appd:more_results}
\begin{figure}[h!]
    \centering
    \includegraphics[width=0.5\textwidth]{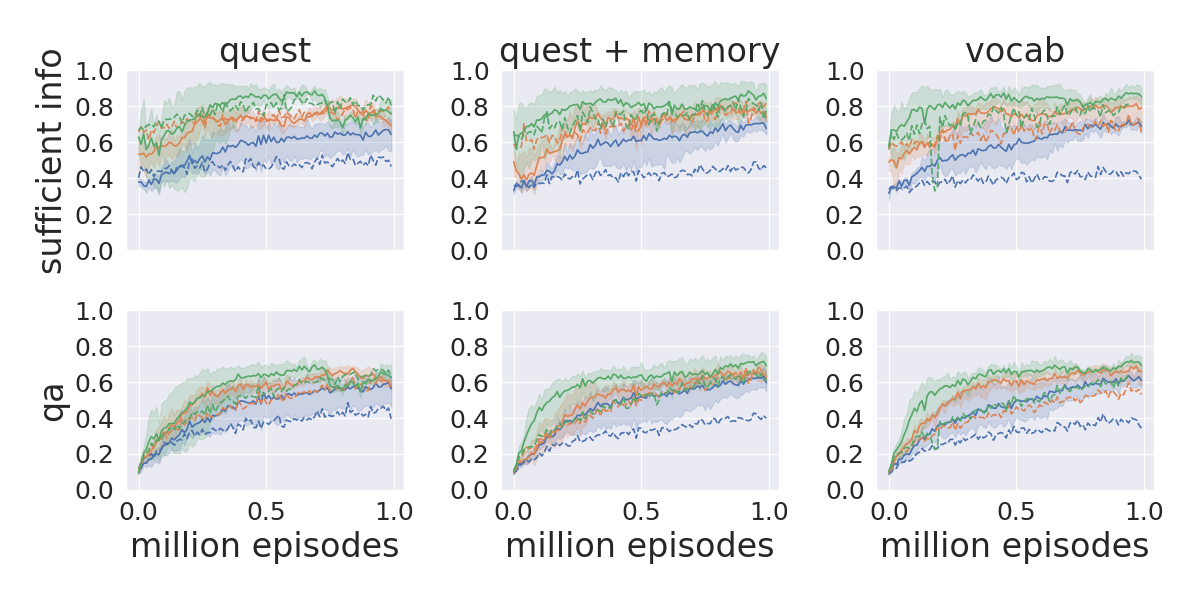}
    \caption{Training performance on \isquad, easy mode. Solid line: \dqn, dashed line: \aac; number of memory slots:  \textcolor{blue1}{1}, \textcolor{orange}{3}, \textcolor{green1}{5}.}
    \label{fig:1}
\end{figure}

\begin{figure}[h!]
    \centering
    \includegraphics[width=0.5\textwidth]{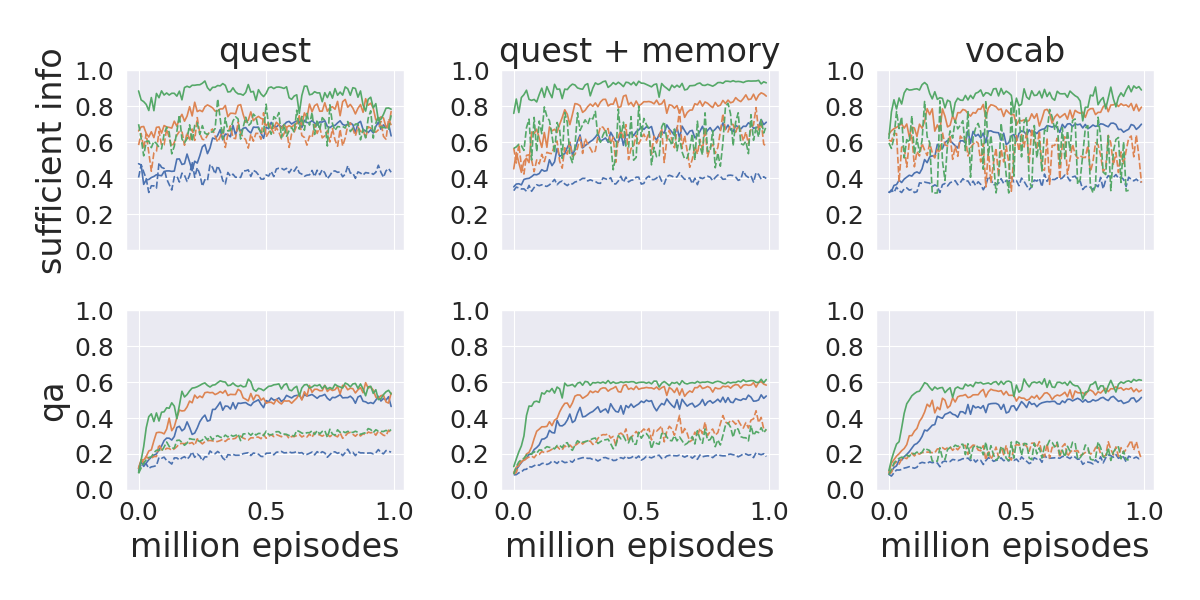}
    \caption{Validation performance on \isquad, easy mode. Solid line: \dqn, dashed line: \aac; number of memory slots:  \textcolor{blue1}{1}, \textcolor{orange}{3}, \textcolor{green1}{5}.}
    \label{fig:2}
\end{figure}

\begin{figure}[h!]
    \centering
    \includegraphics[width=0.5\textwidth]{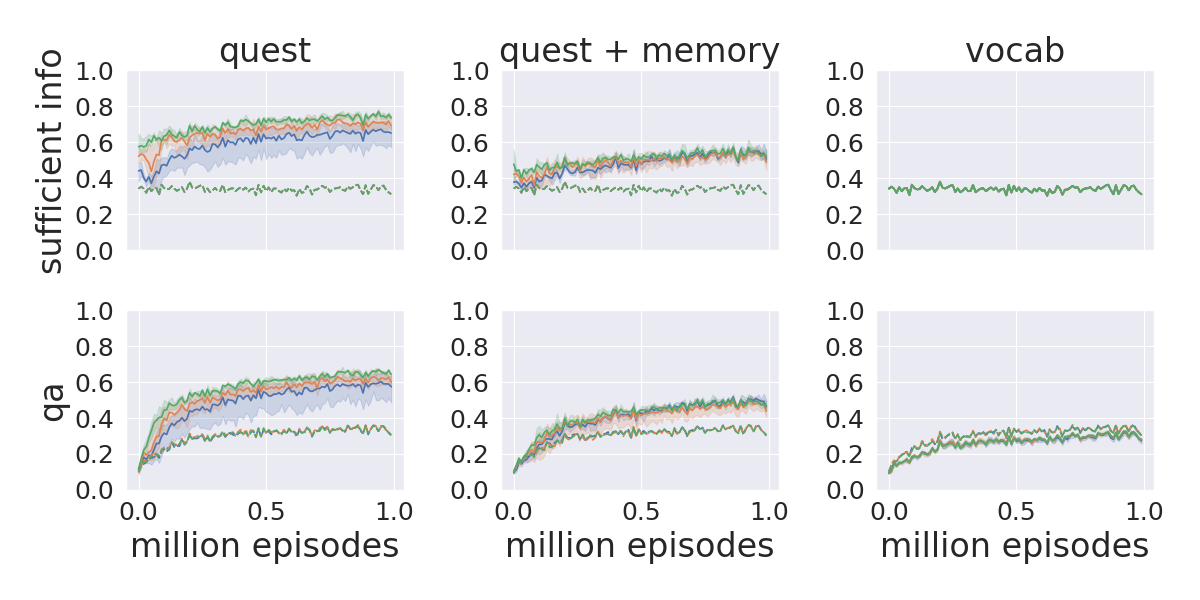}
    \caption{Training performance on \isquad, hard mode. Solid line: \dqn, dashed line: \aac; number of memory slots:  \textcolor{blue1}{1}, \textcolor{orange}{3}, \textcolor{green1}{5}.}
    \label{fig:3}
\end{figure}

\begin{figure}[h!]
    \centering
    \includegraphics[width=0.5\textwidth]{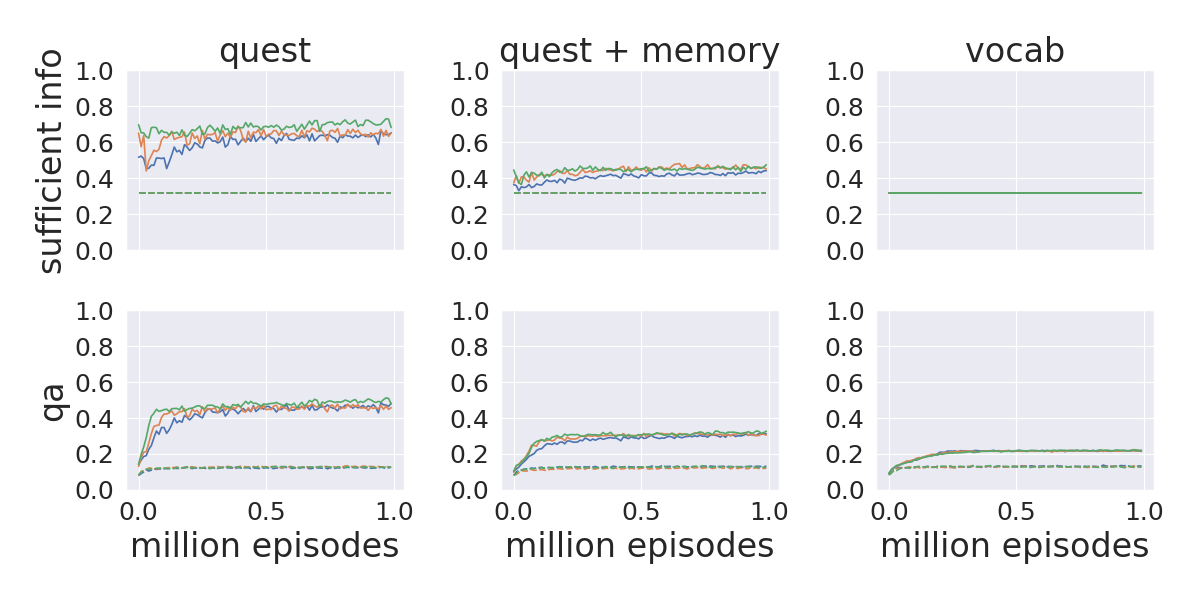}
    \caption{Validation performance on \isquad, hard mode. Solid line: \dqn, dashed line: \aac; number of memory slots:  \textcolor{blue1}{1}, \textcolor{orange}{3}, \textcolor{green1}{5}.}
    \label{fig:4}
\end{figure}

\begin{figure}[h!]
    \centering
    \includegraphics[width=0.5\textwidth]{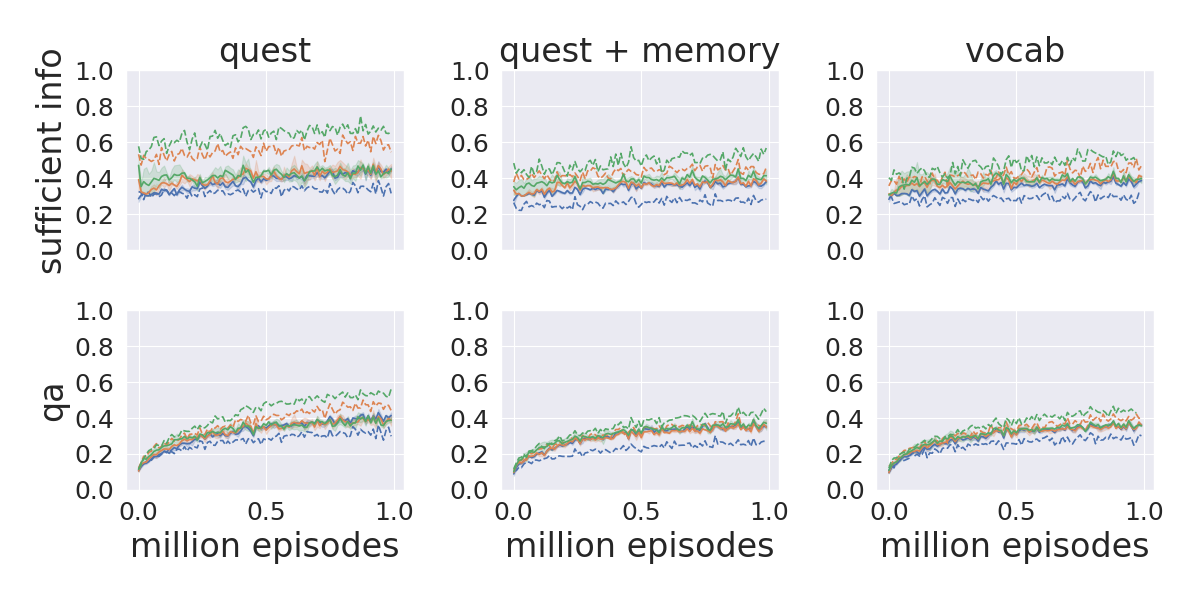}
    \caption{Training performance on \inewsqa, easy mode. Solid line: \dqn, dashed line: \aac; number of memory slots:  \textcolor{blue1}{1}, \textcolor{orange}{3}, \textcolor{green1}{5}.}
    \label{fig:5}
\end{figure}

\begin{figure}[h!]
    \centering
    \includegraphics[width=0.5\textwidth]{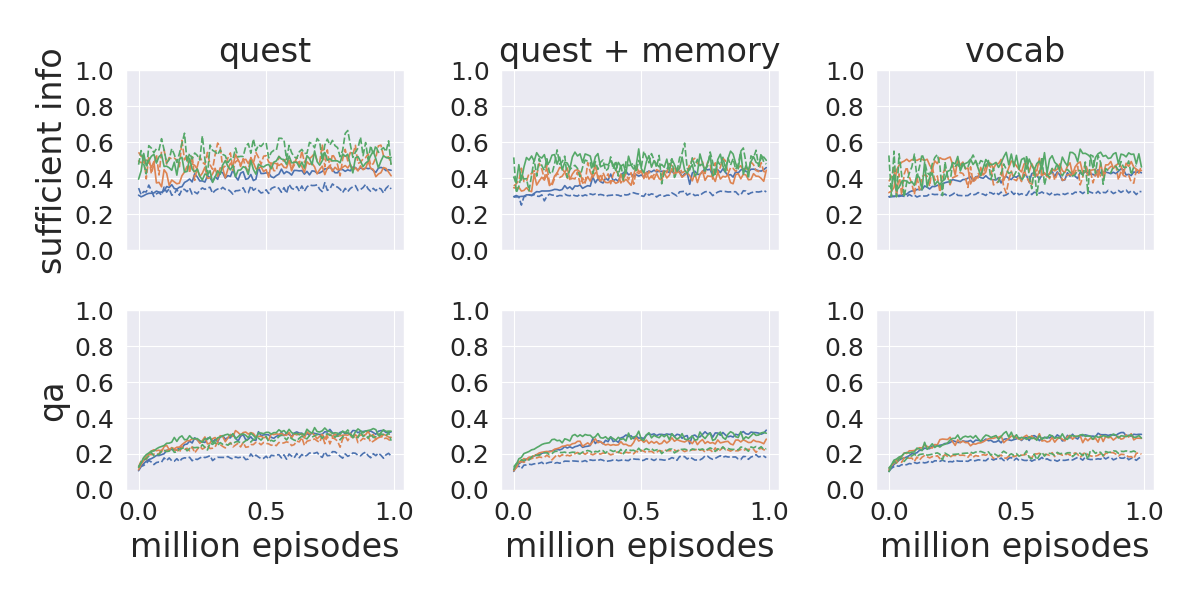}
    \caption{Validation performance on \inewsqa, easy mode. Solid line: \dqn, dashed line: \aac; number of memory slots:  \textcolor{blue1}{1}, \textcolor{orange}{3}, \textcolor{green1}{5}.}
    \label{fig:6}
\end{figure}

\begin{figure}[h!]
    \centering
    \includegraphics[width=0.5\textwidth]{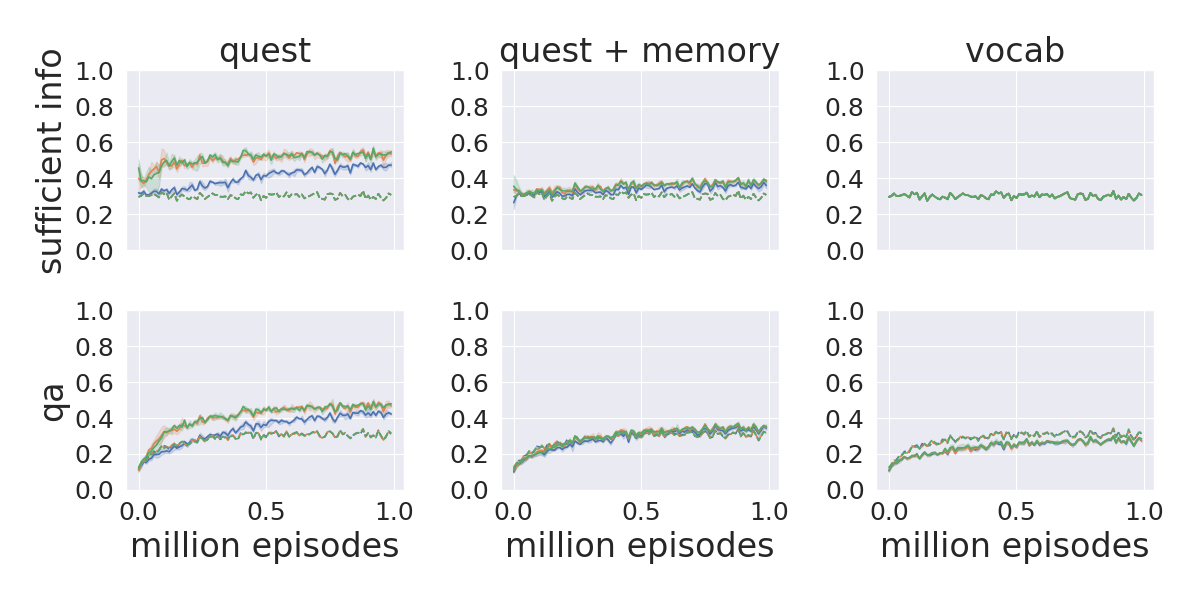}
    \caption{Training performance on \inewsqa, hard mode. Solid line: \dqn, dashed line: \aac; number of memory slots:  \textcolor{blue1}{1}, \textcolor{orange}{3}, \textcolor{green1}{5}.}
    \label{fig:7}
\end{figure}

\begin{figure}[h!]
    \centering
    \includegraphics[width=0.5\textwidth]{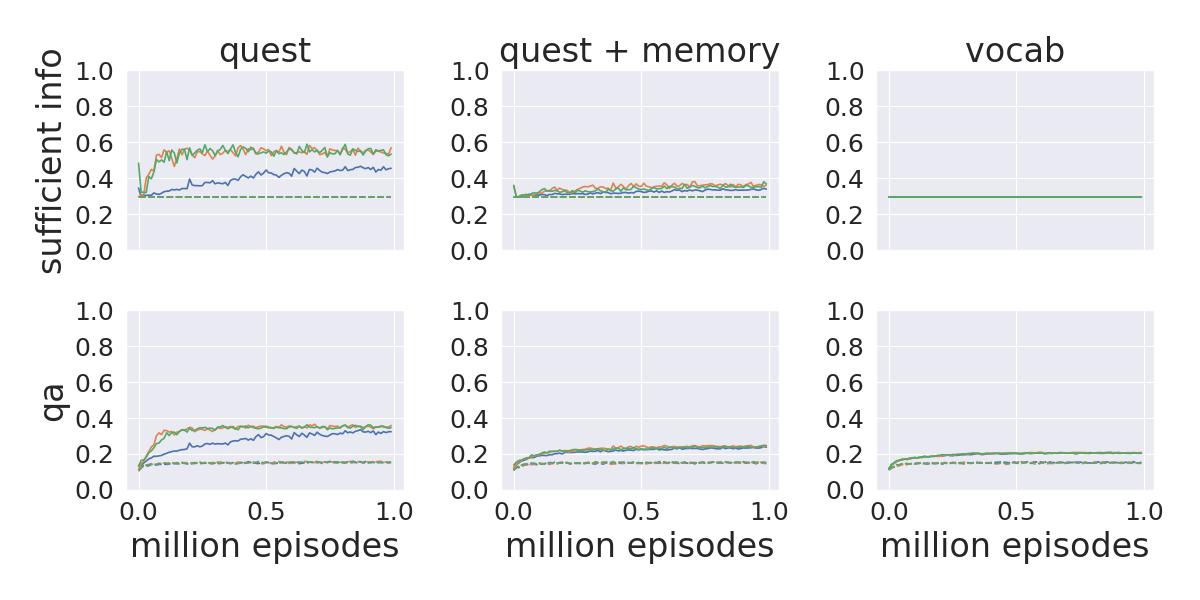}
    \caption{Validation performance on \inewsqa, hard mode. Solid line: \dqn, dashed line: \aac; number of memory slots:  \textcolor{blue1}{1}, \textcolor{orange}{3}, \textcolor{green1}{5}.}
    \label{fig:8}
\end{figure}

\section{Implementation Details}
\label{appd:implementation_details}

In all experiments, we use \emph{Adam} \citep{kingma14adam} as the step rule for optimization, with the learning rate set to 0.00025.
We clip gradient norm at 5.0.
We initialize all word embeddings by the 300-dimensional fastText \citep{mikolov18fasttext} word vectors trained on Common Crawl (600B tokens), they are fixed during training.
We randomly initialize character embeddings by 200-dimensional vectors.
In all transformer blocks, block size is 96.

Dimensionality of $\mlp_{\text{shared}}$ in Eqn.~\ref{eqn:qvalue} is $\mathbb{R}^{96 \times 150}$; dimensionality of $\mlp_{\text{action}}$ is $\mathbb{R}^{150 \times 4}$ and $\mathbb{R}^{150 \times 2}$ in easy mode (4 actions are available) and hard mode (only 2 actions are available), respectively; dimensionality of $\mlp_{\text{query}}$ is $\mathbb{R}^{150 \times V}$ where $V$ denotes vocabulary size of the dataset, as listed in Table~\ref{tab:dataset_stats}.

Dimensionalities of $\mlp_{0}$ and $\mlp_{1}$ in Eqn.~\ref{eqn:qa_mlp01} are both $\mathbb{R}^{192 \times 150}$; dimensionalities of $\mlp_{2}$ and $\mlp_{3}$ in Eqn.~\ref{eqn:qa_mlp23} are both $\mathbb{R}^{150 \times 1}$.

During A2C training, we set the value loss coefficient to be 0.5, we use an entropy regularizer with coefficient of 0.01.
We use a discount $\gamma$ of 0.9 and mini-batch size of 20.

During DQN training, we use a mini-batch of size 20 and push all transitions (observation string, question string, generated command, reward) into a prioritized replay buffer of size 500,000.
We do not compute losses directly using these transitions.
After every 5 game steps, we sample a mini-batch of 64 transitions from the replay buffer, compute loss, and update the network.
we use a discount $\gamma$ of 0.9. 
For noisy nets, we use a $\sigma_0$ of 0.5.
We update target network per 1000 episodes.
For multi-step returns, we sample $n \sim \text{Uniform}[1, 2, 3]$.

When our agent terminates information gathering phase, we push the question answering transitions (observation string at this time, question string, ground-truth answer) into a question answering replay buffer.
After every 5 game steps, we randomly sample a mini-batch of 64 such transitions from the question answering replay buffer and train the model using NLL loss.

For more detail please refer to our open-sourced code.

\end{document}